# HYBRID FUEL CELLS POWER FOR LONG DURATION ROBOT MISSIONS IN FIELD ENVIRONMENTS

Jekanthan Thangavelautham[1], Danielle Gallardo[2],
Daniel Strawser[1], Steven Dubowsky[1]

[1]*Mechanical Engineering Department, Massachusetts Institute of Technology
77 Massachusetts Ave., Cambridge, MA, 02139,
{jekan, dstrawse, dubowsky}@mit.edu*

[2]*Department of Mechanical, Aerospace and Nuclear Engineering, Rensselaer Polytechnic Institute, 110th 8th Jonsson Engineering Center, Troy New York, 12180*

Mobile robots are often needed for long duration missions. These include search rescue, sentry, repair, surveillance and entertainment. Current power supply technology limit walking and climbing robots from many such missions. Internal combustion engines have high noise and emit toxic exhaust while rechargeable batteries have low energy densities and high rates of self-discharge. In theory, fuel cells do not have such limitations. In particular Proton Exchange Membrane (PEMs) can provide very high energy densities, are clean and quiet. However, PEM fuel cells are found to be unreliable due to performance degradation. This can be mitigated by protecting the fuel cell in a fuel-cell battery hybrid configuration using filtering electronics that ensure the fuel cell is isolated from electrical noise and a battery to isolate it from power surges. Simulation results are presented for a HOAP 2 humanoid robot that suggests a fuel cell powered hybrid power supply superior to conventional batteries.

## I. Introduction

Mobile robots, including walking robots are needed to perform long duration missions that are difficult, dangerous and tedious. These include search and rescue, repair, entertainment, sentry, surveillance applications [1, 2]. Continuous operation of these robots, lasting days and weeks (not hours) would be ideal for these applications. Typical power demands for field robots will vary significantly during a mission, often with high peak power demands. These field systems often have constraints on their mass, volume and noise.

Current power supply technology is a key limiting factor for long duration field robotic applications. Internal combustion engines can provide high power for long durations but produce toxic exhaust, noise and strong thermal signatures making them inappropriate for many important applications. Current rechargeable batteries have very low energy densities and high rates of self-discharge, requiring systems to stop and recharge every few hours, making them ineffective for continuous long duration missions. Hence, there is a significant need for a power supply that can provide the high total energy required for long duration missions that is quiet and clean.

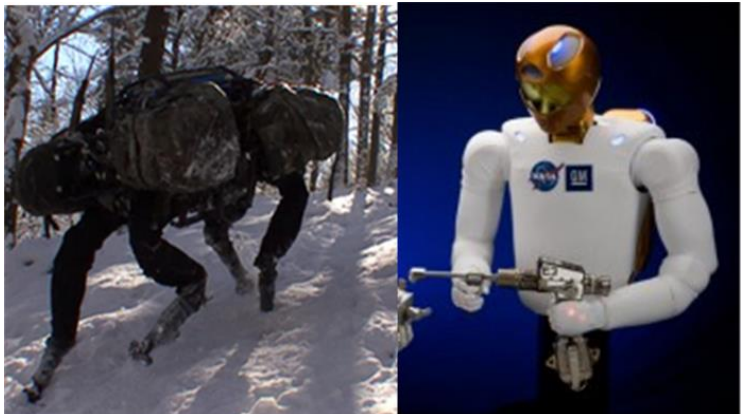

**Figure 1:** (Left) Boston Dynamics Big Dog, a four-legged supply robot. (Right) Robonaut 2 repair robot.

## II. Fuel Cell Power for Mobile Robots

Fuel cells are high energy sources of power that have been suggested for robots [7, 8]. They are a promising alternative mobile source of power and have the potential to overcome limitations of current batteries and internal combustion engines. They are simple electrochemical devices that convert chemical energy into electricity (Figure 2). Unlike a battery, fuel cells require a constant supply of fuel and oxidant to produce electricity. Proton Exchange Membrane (PEM) fuel cells are particularly attractive for robotics. These devices consist of simple solid state components sandwiched together as shown in Figure 2. They combine hydrogen fuel, and oxygen (from breathing air) through the most energy releasing reaction known, to produce electricity and water. It has been demonstrated; PEM fuel cells can reach 65-70 % or higher operating efficiencies at room temperature and produce clean water exhaust [9].

## III. The Challenges of Fuel Cells for Robots

While PEM fuel cells are simple and sound great in theory, they have three fundamental problems for practical robotics applications. These problems are storage of hydrogen fuel, long-life reliability of fuel cells and low power. Hydrogen fuel due to its high energy content and low density is difficult to store. In our research, we have developed simple, innovative hydrogen storage technologies that promise energy storage densities better than the best batteries of today [5].

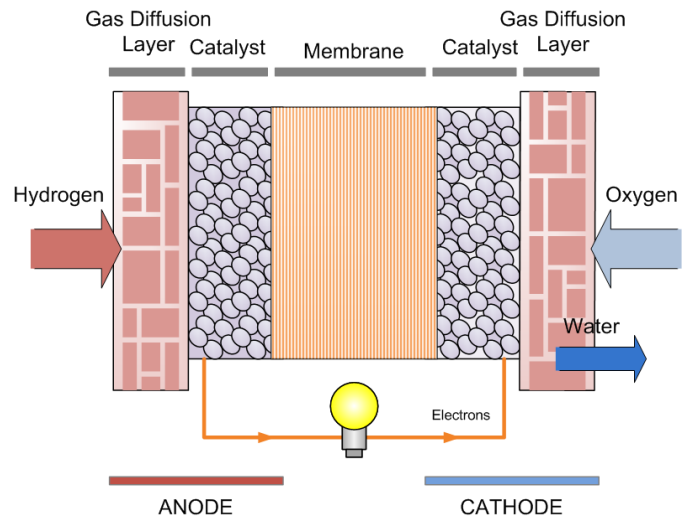

**Figure 2:** A PEM Fuel Cell Consumes Reactants, Hydrogen and Oxygen to Produce Electricity, Water and Heat.

Second, PEM fuel cells have been found to be unreliable [3]. Our studies of PEM fuel cells show that they are delicate and unreliable due to degradation of their components, resulting in short lives and premature failure [4]. However our physical models and experiments suggests that PEM fuel cells controlled to operate within narrow operating can be made robust, have long lives of 3-5 years or more and high operating efficiencies [4]. Among the factors known to degrade fuel cells, are high operating voltages and electrical noise. As discussed below, mobile field robots operating in unstructured environments are subject to very substantial variation, which, without proper control can result in fuel cell degradation that shortens their lives. A solution to this problem is discussed in the section below.

A third problem with fuel cells is that while they are high energy devices, they have relatively low power. This is a problem for robotics, where typical power requirements can vary substantially over a mission, with low-power rest periods and short bursts at peak power. These varying power demands are known to stress the fuel cells, resulting in short lives. A solution is to use fuel-cells in a hybrid system for mobile robots that maintains a fuel cell at optimal operating conditions to maximize life and efficiency, by protecting it from external electrical load variations, noises and meeting peak power requirement using a battery (see Figure 3).

Fuel cell hybrid systems have been subjected to meet rapid, transient power demands in large and stationary applications, and in robotics, to meet power surges [7, 8]. However these hybrid system designs have not considered the effects of fuel cell degradation.

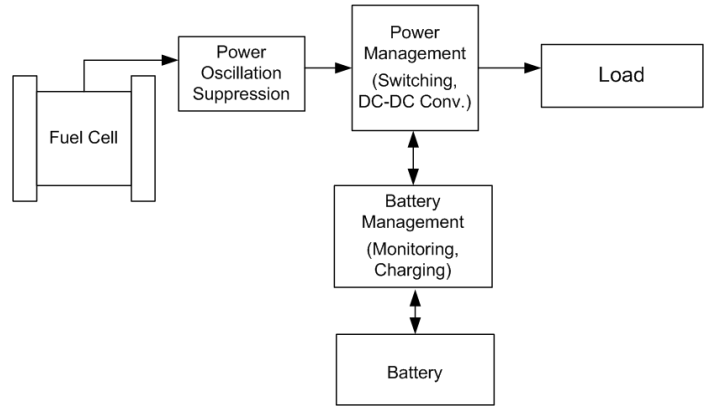

**Figure 3:** Proposed Fuel Cell-Battery Hybrid Power Supply for Robots.

## IV. Research

The research presented here is focused on developing hybrid system design concept for mobile robots that have energy densities that exceed the best battery technology. The hybrid system is designed to meet the required peak power demands and isolate the fuel cell from degrading stresses of high and low frequency noises generated by conditioning circuits required for battery management. Physical models are used to simulate expected conditions and control systems are developed to demonstrate the concept. It is shown that the results are vast improvement over conventional batteries in terms of life, efficiency, energy density and power density.

## V. Case Study: Power for Humanoid Walking Robot

Here a hybrid fuel cell power supply for a HOAP-2 humanoid walking robot (Figure 4) developed by Fujitsu is presented. The HOAP-2 is a 7.8 kg robot, with a maximum rated power of 250 W. It contains a 1.2 kg Nickel Metal Hydride rechargeable battery pack by default. The system contains 25 servo actuators, 6 for each leg and 5 for each arm, 2 for the head and 1 one for the waist. The robot has onboard computer equivalent to a PC-104 Pentium III system, a vision system consisting of 2 CCD cameras, onboard accelerometers, gyroscope and pressure sensors on each feet.

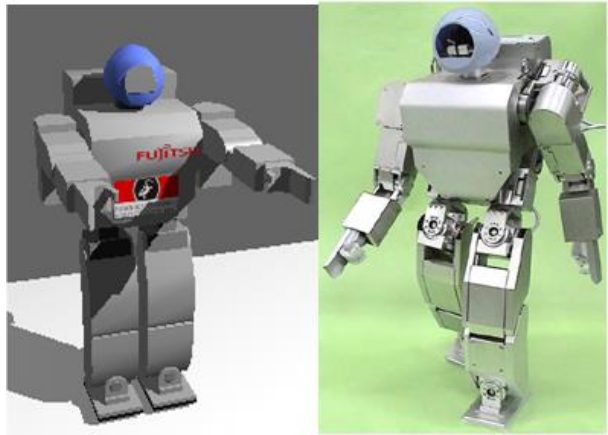

**Figure 4:** (Right) Fujitsu's HOAP 2 Robot (Left) Cyberbotics Webots[TM] Model of the HOAP-2.

A simulation model of HOAP-2 from Cyberbotics Webots[TM] [6] is used for our power demand calculations. The robot system consists of three different subsystems for power calculations, namely electro-mechanical system, computer + sensors and power system. For the electro-mechanical system the simulator model provides mechanical power output of the servo motors. The servo motors are assumed to have a 50 % electrical to mechanical efficiency. The computer and sensor system are assumed to be always powered and consume 40 W based on the HOAP-2 specifications. Below, power demand profiles of the robot's walking behavior is shown (Figure 5). For these scenarios, alternative power sources are compared with the default nickel metal hydride battery packs that weigh 1.2 kg.

**VI. Fuel Cell Hybrid System**

The fuel cell hybrid system consists of a fuel cell stack that provides steady power source and rechargeable lithium ion A123 Nanophosphate[TM] battery that meets peak power demands. The fuel cells within the stack are operated at constant operating voltage of 0.8, providing a 65 % operating efficiency. Our research into the degradation of fuel cells, based on models and experimental results shows that increased operating voltages exponentially decreases the life of the fuel cell [4]. Operating the fuel cell at constant voltage of 0.8 V or less ensures long-life, while providing sufficiently high operating efficiency. The fuel cell trickle charges the battery during idle times, ensuring the battery is fully charged, to meet power peaks. The Nanophosphate[TM] battery handles peak demands and can better handle deep battery discharges compared to conventional lithium ion batteries. Ensuring the battery is nearly fully charged, maximizes its life. The battery for the hybrid system is sized based on its specific power density to meet the maximum possible power requirements of the robot.

An oscillation suppression circuit interfaces a fuel cell to the power management system, consisting of power switching circuitry and a DC-DC convertor. The interface circuit effectively extracts the energy from the fuel cell and transfers it into the battery. The oscillation suppression circuits prevents any voltage oscillation from electrical circuits, particularly DC-DC convertors from being noticed by the fuel cell. This ensures the fuel cell operates at steady operating voltage, without any electrical load oscillations.

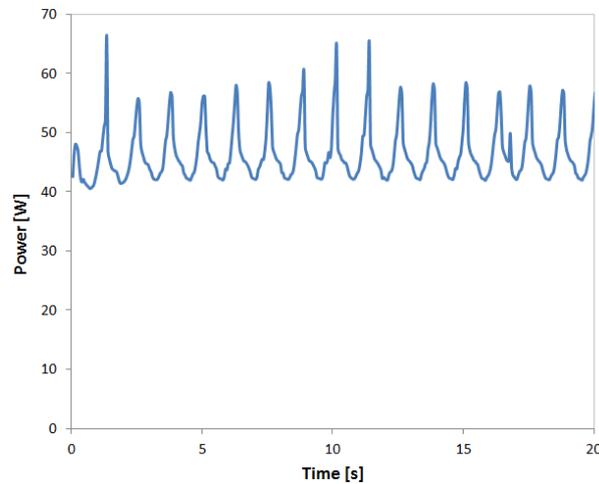

**Figure 5:** Power Demand of a HOAP-2 Robot, walking at 0.06 m/s.

**VII. Hybrid System Sizing**

Based on the power demand profiles (Figure 5), the fuel cell will provide a constant steady source of power of 45 W. While the power peaks will be handled by the battery. For 45 W steady supply of power, a fuel cell stack weighing 150 g will be required. For a 250 W peak required by the robot, a 135 g lithium ion Nanophosphate$^{TM}$ battery is required (with specific power of 1,850 W/kg for APR18650 cells) . Another 115 g is allocated for mass of power electronics and other items, leaving 800 g for the Lithium Hydride fuel supply with an energy density of 4,950 Wh/kg.

**VIII. Power System Comparison**

Four power supply configurations are compared, including a nickel metal hydride and lithium ion battery system, a fuel cell system and a fuel cell hybrid system. See Table 1.

**Table 1:** Power Supply Comparison for HOAP-2 Humanoid Robot

| Power Supply | FC Stack Mass | FC Fuel Mass | Energy Density | System Life | Run-Time |
|---|---|---|---|---|---|
| NiMH Battery | - | - | 40 Wh/kg | 0.3 year | 3 hours |
| Li Ion Battery | - | - | 120 Wh/kg | 1 year | 9 hours |
| Fuel Cell | 300 g | 900 g | 4950 Wh/kg | 5 days | 99 hours |
| Fuel Cell Hybrid | 150 g | 800 g | 4950 Wh/kg | 3 years | 88 hours |

Nickel metal hydride and lithium ion batteries have the lowest energy densities and thus provide short run-times, before requiring recharging. The system life of the batteries is computed based on expected lifetime of 1,000 charge/recharge multiplied by the run-times hours. A direct fuel cell system has the longest run-time. However the life of the system, based on our degradation models is expected to last just 5 days making this option impractical. The fuel cell hybrid system offers a good trade-off between both run-time and system life.

**IX. Summary and Conclusions**

Based on these results, the fuel cell hybrid system concept offers high energy density, long-life and would meet the required peak power demands of a battery. The key to our hybrid system concept is effective control and design, where a fuel cell and battery are optimally sized to minimize stress on the fuel cell, while enabling a battery to meet power demand peaks. By minimizing stresses on the fuel cell, the system can be operated for long-lives at high operating efficiencies.